\documentclass[11pt]{article}

\usepackage[english]{babel}

\usepackage[
  backend=biber,
  style=vancouver,
  sorting=none
]{biblatex}
\usepackage{amsmath,amssymb,amsfonts}
\usepackage{algorithm} 
\usepackage{algpseudocode}
\usepackage{graphicx}
\usepackage{caption}
\usepackage{appendix}
\usepackage{geometry}
\usepackage{kotex}
\usepackage{authblk}
\usepackage{xcolor}
\usepackage{hyperref}
\usepackage{booktabs} % for nice tables
\usepackage{arydshln}
\usepackage{graphicx}
\usepackage{gensymb}
\usepackage{multirow}
\hypersetup{
    colorlinks=False,
    linkbordercolor=red,
}
\usepackage{multirow}

\newcommand\equalcontribution{\thanks{These authors contributed equally to this work.}}
\newcommand\correspondingauthor{\thanks{Corresponding author. \\ \textit{e-mail:} ijkim@handong.edu}}
\begin{document}

\pagenumbering{gobble}
%TC:ignore
\title{ARNAI: Artifact Removal Network based on Autoencoding and Inpainting for Robust Spinal Image Segmentation and Measurement}

% \author[1]{{Anonymous Authors}}
\author[1]{Sang-Jin Park}
\author[3]{Jinyoung Choi}
\author[3]{Seokwon Kim\equalcontribution}
\author[3]{Seungeon Song\protect\footnotemark[1]}
\author[3]{Insu Park\protect\footnotemark[1]}
% \fntext[fn1]{These authors contributed equally to this work.}
\author[1,2]{Dougho Park}
\author[1]{Taeyeon Kim}
\author[1]{Youjin Lee}
\author[1]{Donghoon Yang}
\author[1]{Jaeman Cho}
\author[1]{Joongwon Yang}
\author[1]{Mansu Kim}
\author[1]{Heumdai Kwon}
\author[4]{Hong Gyu Baek}
\author[4]{Dae Chul Cho}
\author[3]{Injung Kim\correspondingauthor}    % modified by ijkim 20240308

\affil[1]{Pohang Stroke and Spine Hospital, Pohang, Republic of Korea}
\affil[2]{School of Convergence Science and Technology,
Pohang University of Science and Technology, Pohang, Republic of
Korea }
\affil[3]{School of CSEE, Handong Global University
, Pohang, Republic of Korea}
\affil[4]{Department of Neurosurgery, Kyungpook National University Hospital, Kyungpook National University College of Medicine, Daegu, Republic of Korea}

\date{}
\renewcommand*{\bibopenbracket}{(}
\renewcommand*{\bibclosebracket}{)}
\DeclareFieldFormat{labelnumber}{#1}            % 실제 라벨
\DeclareFieldFormat{labelnumberwidth}{#1\adddot}% 매단 들여쓰기 폭 + 점\begin{document}
\maketitle

\section*{Key Points}
\begin{enumerate}
    \item A novel artifact removal model ARNAI was introduced to address the challenge in spinal radiograph segmentation when lumbar vertebrae are obscured by spinal implants.

    \item In experiments, ARNAI mitigated implant-related artifacts in implant-containing radiographs and improved segmentation accuracy, increasing the Dice similarity coefficient (DSC) to 0.870 from 0.814 and reducing L4–L5 segmental Cobb angle error by approximately 70\%.

     \item The \textit{Restore, Segment, and Measure} (RSM) framework incorporating ARNAI substantially improved automated spinal radiograph segmentation and enhanced downstream spinopelvic parameter measurement, including pelvic tilt, sacral slope, lumbar lordosis, and L4–L5 segmental Cobb angle.

 \end{enumerate}

\section*{Abbreviation}
RSM = Restore, Segment, and Measure, ARNAI = Artifact Removal Network based on Autoencoding and Inpainting, VQ-GAN = Vector Quantized Generative Adversarial Network, DSC = Dice Similarity Coefficient, PT = Pelvic Tilt, SS = Sacral Slope, LL = Lumbar Lordosis,  SCA = Segmental Cobb Angle, ICC = Intraclass Correlation Coefficient

\section*{Summary} % no more than 255 characters in boldface
\textbf{
We propose a Restore, Segment, and Measure framework that reduces implant-related artifacts in postoperative spinal radiographs using autoencoding and inpainting without implant-containing training data, then segments the lumbar spine and measures spinopelvic parameters. }

\section*{Keywords}
Medical image restoration, Medical image segmentation, Spinopelvic parameter estimation, Generative models
\newpage
\begin{abstract}

\textbf{Purpose}: This study aims to develop an AI framework applicable for postoperative imaging for automated measurement of spinopelvic parameters on radiographs with robustness to the presence of spinal implants.

\textbf{Materials and Methods}: We retrospectively reviewed lateral lumbar spine radiographs from two institutions (Internal: January 2017--December 2024; External: October 2021--September 2025). We developed the \textit{Restore, Segment, and Measure} (RSM) framework, incorporating a novel \textit{Artifact Removal Network based on Autoencoding and Inpainting} (ARNAI) to mitigate implant-related artifacts in postoperative radiographs. Segmentation and spinopelvic parameter (PT, LL, SS, SCA) measurement performance were assessed using Wilcoxon signed-rank tests and intraclass correlation coefficients.

\textbf{Results}: When ARNAI was added to a recent Transformer-based segmentation model, FCBFormer, the mean DSC increased to 0.870 from 0.814, with marked gains at L3–L5 and smaller improvements at L1–L2. On 91 radiographs with implants, the mean L4–L5 segmental Cobb angle error decreased to $4.7^\circ$ from $15.6 \sim 16.2^\circ$, an average error reduction of 70\%. The ICC for L4–L5 segmental Cobb angle improved to 0.54 (Rater 1) and 0.59 (Rater 2) from 0.18, and ICCs for pelvic tilt, lumbar lordosis, and sacral slope all exceeded 0.70. The improvement in L4–L5 segmental Cobb angle error was statistically significant in the internal implant-containing cohort after correction for multiple comparisons.

\textbf{Conclusion}: 
The proposed RSM framework improved automated spinopelvic parameter measurement in implant-containing postoperative radiographs. By mitigating implant-related artifacts, ARNAI improved segmentation and downstream measurement accuracy, with the greatest benefit observed for L4–L5 segmental Cobb angle estimation, where the mean error was reduced by approximately 70\%. 
 \end{abstract}
\newpage
%TC:endignore

\section{Introduction}
    Spinal sagittal alignment is closely related to pain, neurologic symptoms, and quality of life in patients with adult spinal deformity \cite{Three-dimensional-analysis, Sagittal-balance-of-the-spine, Scoliosis-Research-society, Radiographical-spinopelvic-parameters, Sagittal-balance-and-health-related}. In clinical practice, spinopelvic parameters are essential for evaluating sagittal balance, postoperative correction, and longitudinal follow-up.  Pelvic tilt (PT), sacral slope (SS), and lumbar lordosis (LL) are commonly measured on lateral lumbar radiographs, but manual measurement is time-consuming and subject to interobserver variability. 

Deep learning approaches have been proposed to automate spinopelvic assessment and have shown high performance on radiographs without implants~\cite{Spino-Unet, Spino-MultiResUNet, Spino-MaskRCNN}. However, their generalizability is substantially compromised in postoperative cases, where metallic artifacts interfere with contour detection~\cite{Keypoint2024, You25}. This limitation is clinically important because lateral lumbar radiographs are routinely used after lumbar fusion to assess postoperative radiologic outcomes, including sagittal alignment. Prior strategies to mitigate this are limited: some exclude implant-containing images, restricting clinical applicability \cite{Spino-MaskRCNN}, whereas others depend on scarce annotated implant-containing datasets or expert-labeled keypoints \cite{Spino-MultiResUNet, Keypoint2019, Keypoint2021, Keypoint2024}.

To address these limitations, we propose the \textit{Restore, Segment, and Measure} (RSM) framework, which incorporates the \textit{Artifact Removal Network based on Autoencoding and Inpainting} (ARNAI) for robust spinopelvic parameter estimation in implant-containing radiographs. ARNAI uses a Vector Quantized Generative Adversarial Network (VQ-GAN) \cite{VQGAN} trained exclusively on implant-free images to localize implant-related artifacts and mitigate their effects without requiring expert-labeled implant-containing training data. The inpainted radiographs are then segmented to measure PT, SS, LL, and L4–L5 segmental Cobb angle (SCA); L4–L5 was selected for its clinical relevance, common involvement in degenerative lumbar pathology, and frequent inclusion in lumbar fusion surgery. This framework highlights the potential of generative artifact removal for automated postoperative spinal analysis.

\section{Materials and Methods}

\subsection{Data Acquisition and Preparation}

This retrospective study used lateral lumbar spine radiographs collected from two institutions. The internal cohort was retrieved from the institutional imaging archive between January 2017 and December 2024 and included 2,486 radiographs from 1,683 patients, comprising both non-instrumented radiographs and postoperative radiographs after lumbar fusion surgery. Because the internal cohort was retrospectively assembled from a historical archive without a prospectively logged screening workflow, criterion-specific exclusion counts were not recoverable. 

For external validation, postoperative radiographs from an independent tertiary referral hospital were retrieved between October 2021 and September 2025. Of 443 postoperative radiographs reviewed for eligibility, 226 were excluded according to predefined criteria, including inadequate field-of-view or insufficient visualization of key anatomical landmarks. The remaining 217 implant-containing radiographs were used for external testing and expert reference assessment. The external dataset contains more challenging samples with lower image quality and non-standard acquisition postures. Implant-free internal samples were divided into training, validation, and test sets in an 8:1:1 ratio, whereas implant-containing samples in both datasets were used only for testing. Baseline characteristics and imaging details are summarized in Supplementary Tables 1 and 2, and detailed exclusion criteria are provided in Supplementary Material A.1.2. 

Expert annotations included the L1–L5 vertebral bodies, sacrum, and both femoral heads, and were used for landmark extraction and measurement of PT, SS, LL, and L4–L5 SCA. Detailed annotation, adjudication, and quality-control procedures are provided in the Supplementary Material A.1.3. Institutional review board approval was obtained at both institutions, and informed consent was waived because of the retrospective design and anonymized data.

\subsection{Model Development}

\begin{figure}[h]
\centering
\includegraphics[width=\textwidth]{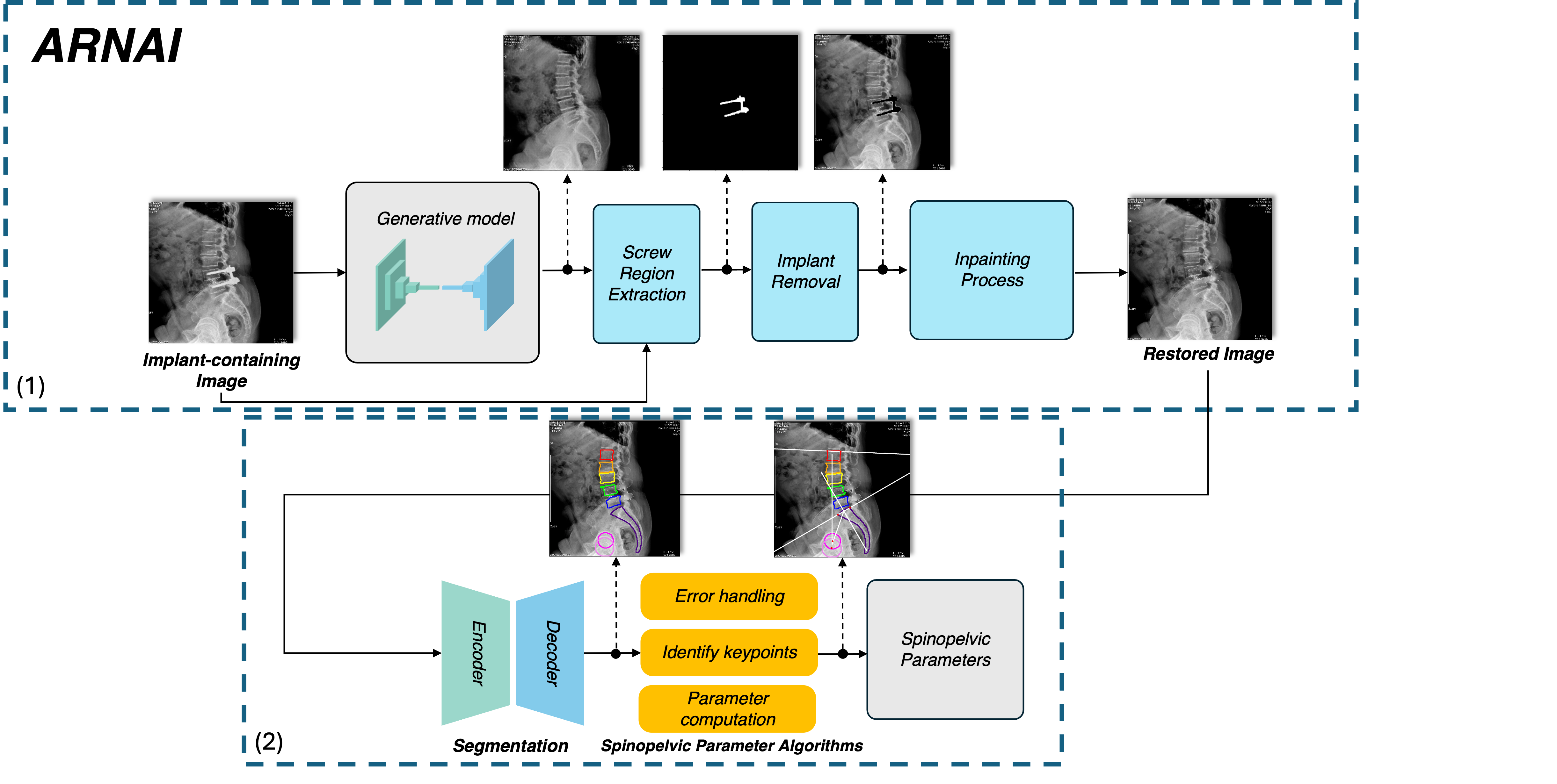}
\captionsetup{font=footnotesize}
\caption{An overview of the RSM framework. (1) Mitigation of implant-related artifacts using ARNAI, and (2) Segmentation mask extraction and spinopelvic parameter estimation results.}\label{fig:arc}
\end{figure}

Fig.~\ref{fig:arc} illustrates the two-stage RSM framework: ARNAI, which mitigates implant-related artifacts in postoperative radiographs, followed by downstream segmentation and spinopelvic parameter estimation. The mathematical formulation and algorithmic details are provided in Supplementary Materials A.2 and A.5.

\begin{figure}[htb!]
\centering
\includegraphics[width=0.9\textwidth]{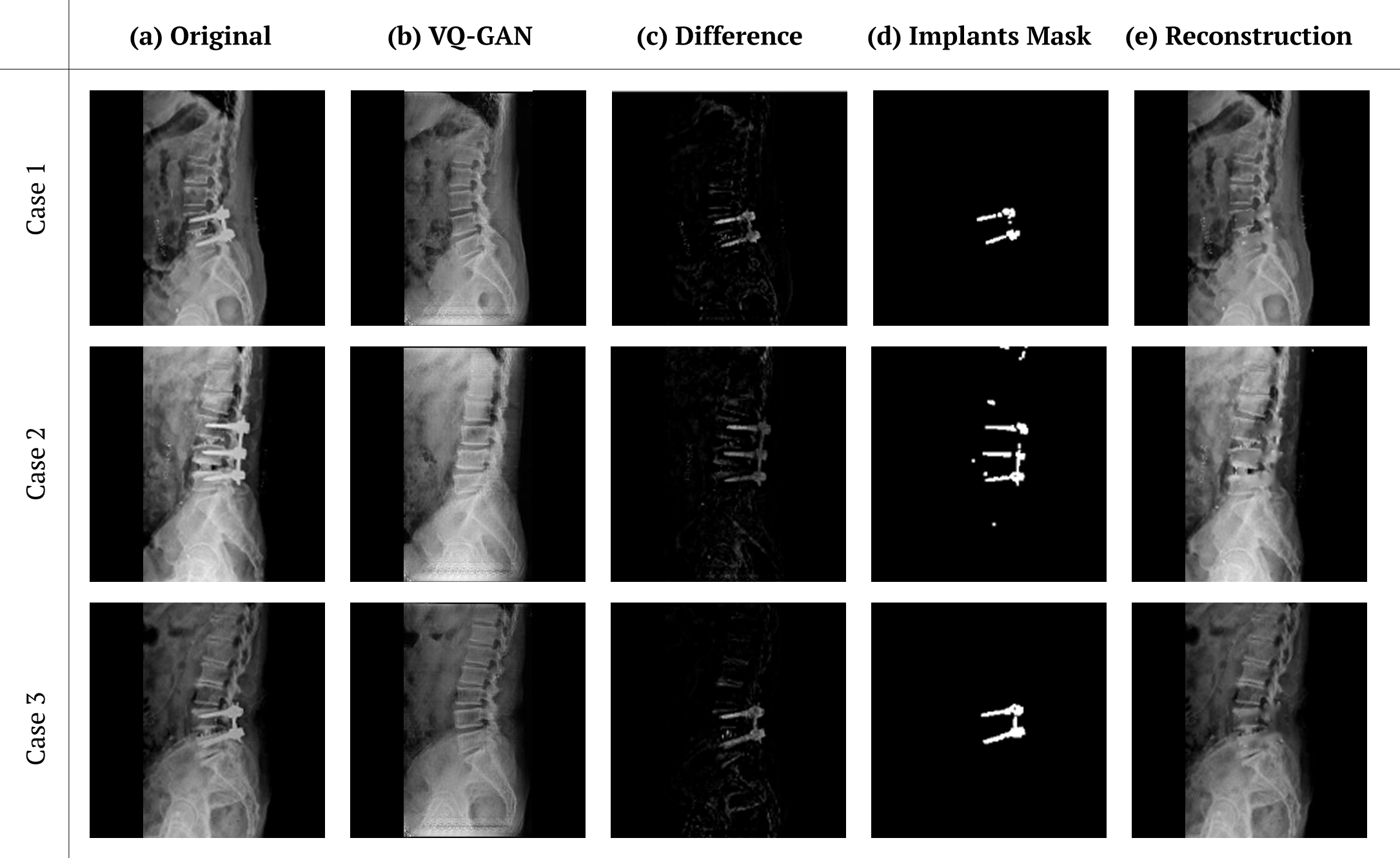}
\captionsetup{font=footnotesize}
\caption{The procedure of ARNAI. (a) Gamma-corrected original input image. (b) The output of VQ-GAN. (c) The intensity difference between (a) and (b). The implant regions exhibit higher values than other regions. (d) The binary mask of implant regions. (e) The reconstructed image in which the implant regions were removed and reconstructed by the inpainting algorithm.}
\label{fig:sc_process}
\end{figure}

\paragraph{Artifact Removal}
Spinal implants introduced during lumbar fusion procedures often obscure anatomy on radiographs and impair segmentation accuracy. To mitigate this, we developed a generative discrepancy-based implant localization and removal framework without requiring labeled implant data for training, as demonstrated in Fig.~\ref{fig:sc_process}.

First, the postoperative radiograph is processed using a VQ-GAN autoencoder \cite{VQGAN} trained exclusively on implant-free data. Due to their out-of-distribution nature, implants are suppressed in the reconstructed output (Fig.~\ref{fig:sc_process}(b)). Because anatomical details outside the implant regions are not faithfully preserved, the VQ-GAN output is used only for implant localization and not for downstream segmentation or measurement.

\begin{figure}[h]
\centering
\includegraphics[width=0.7\textwidth]{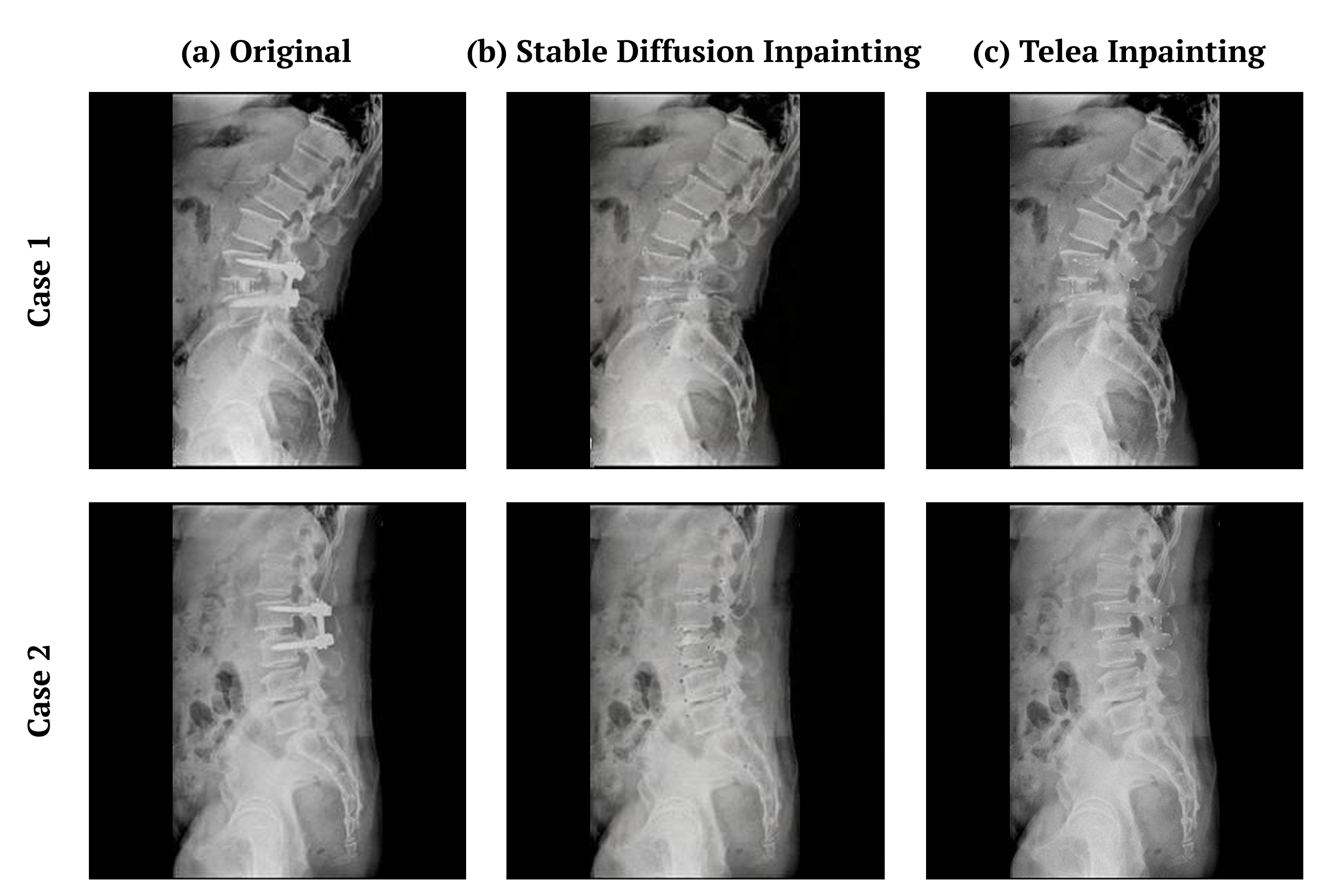}
\captionsetup{font=footnotesize}
\caption{
Qualitative comparison of inpainting methods for implant removal in spine radiographs. The figure presents two different cases (rows). (a) Original radiographs showing implants in the spine. (b) Inpainting results using Stable Diffusion~\cite{stablediffusion}. (c) Inpainting results using Telea inpainting algorithm. As discussed in the text, Telea inpainting algorithm (c) better preserves the original structure of the bones and introduces fewer artifacts compared to Stable Diffusion Inpainting (b).
}
\label{fig:compare_inpaint}
\end{figure}

Implant regions in the reconstructed image exhibit substantial intensity discrepancies compared with the input radiograph, whereas non-implant regions remain largely consistent. Implant regions are thus identified from the pixel-wise difference between the two images: After gamma correction, a difference map is computed, and binarized using Otsu thresholding, followed by morphological opening and dilation for noise removal.
Subsequently, the identified implant regions are erased, and inpainting is applied to reconstruct the missing areas. Instead of recent diffusion–based inpainting~\cite{stablediffusion}, we employ the Telea algorithm \cite{telea} because it better preserves non-implant regions, as compared in Fig.~\ref{fig:compare_inpaint}.

\paragraph{Segmentation and Spinopelvic Parameter Measurement}
The restored radiograph is processed by a segmentation model to extract masks of the lumbar vertebrae (L1–L5), sacrum, and bilateral femoral heads (F1, F2). Among several candidates that are publicly available, FCBFormer \cite{FCBFormer} was selected based on preliminary experiments due to its consistently higher DSC across multiple structures and stable performance on post-operative images.

Spinopelvic parameters (PT, SS, and LL) were measured on lateral lumbar radiographs using predefined anatomical landmarks derived from segmentation outputs via computer vision algorithms, including contour tracing and corner detection. These landmarks included vertebral and sacral endplate points as well as femoral head centers, according to standard radiographic conventions. The segmental Cobb angle at L4–L5 was calculated as the angle between the superior endplate of L4 and the inferior endplate of L5.

\subsection{Performance Evaluation and Statistical Analysis}

%\paragraph{Evaluation of Segmentation Performance}
Segmentation performance was evaluated using the Dice similarity coefficient (DSC) between predicted and ground-truth (GT) masks for the lumbar vertebrae (L1–L5), sacrum, and bilateral femoral heads, averaged across the samples.

%\paragraph{Statistical Analysis of Parameter Measurement}
Accuracy of spinopelvic parameter measurement was evaluated by comparing model predictions with expert reference measurements. Inter- and intra-rater reliability of the reference measurements were assessed using intraclass correlation coefficients based on two-way random-effects [ICC(A,1)] and mixed-effects [ICC(3,1)] models, respectively. Intra-rater reliability was evaluated in 50 randomly selected radiographs re-measured by a board-certified spinal neurosurgeon two weeks after the initial assessment. Inter- and intra-rater agreement were further evaluated using Bland–Altman analysis, including the mean difference, mean absolute difference, and 95\% limits of agreement.

Per-image absolute errors between the baseline and RSM pipelines were compared using two-sided Wilcoxon signed-rank tests on the common subset measurable by both models and raters. Multiple comparisons across four spinopelvic parameters and two raters were corrected using the Benjamini–Hochberg false discovery rate procedure within each cohort, with adjusted $q$-values reported alongside raw $p$-values; $q < 0.05$ was considered significant.

\section{Results}

\subsection{Segmentation Performance}
\begin{figure}[t]
\centering
\includegraphics[width=0.9\textwidth]{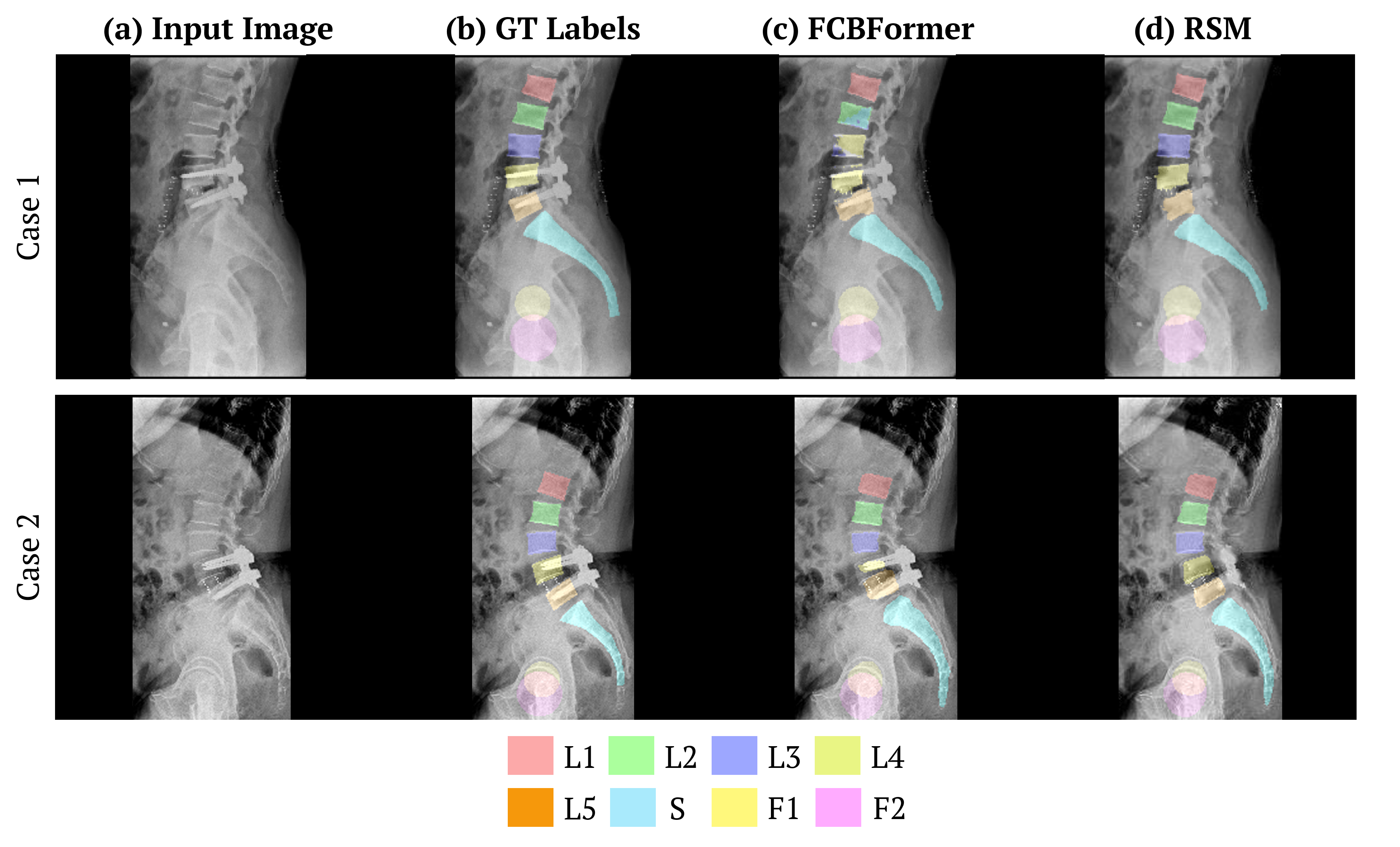}
\captionsetup{font=footnotesize}
\caption{
Qualitative comparison of segmentation results with and without the proposed ARNAI module on two implant-containing spine radiographs (cases 1 and  2). From left to right: (a) original input image, (b) ground-truth annotation, (c) baseline FCBFormer output (DSC ≈ 0.7651/0.8093), and (d) RSM output (DSC ≈ 0.9265/0.9119). Color overlays denote vertebrae L1–L5, sacrum (S), and bilateral femoral heads F1 and F2. Incorporation of ARNAI mitigates implant-related artifacts and improves anatomical boundary delineation. }
\label{fig:results_visualization}
\end{figure}

As summarized in Table~\ref{tab:before_after_ARNAI}, incorporating ARNAI yielded consistent improvements in DSC across all lumbar levels on the internal test set, increasing the overall mean DSC to 0.8697 from 0.8136. Specifically, substantial DSC gains were observed for L3 (+0.096), L4 (+0.184), and L5 (+0.136), which are most frequently affected by implants. However, DSC values for sacrum and F1 were slightly decreased.
On the external test set with domain shift and lower image quality, both models showed reduced DSC values, but RSM remained more robust, especially at L4 and L5, where implants were most frequent (+0.1677 and +0.1448, respectively).
Fig.~\ref{fig:results_visualization} demonstrates qualitative improvements. The baseline model incorrectly segmented the L3 and L4 regions in Case 1 and the L4 and L5 regions in Case 2, whereas RSM accurately restored and segmented these regions.

\subsection{Spinopelvic Parameter Estimation}

Parameter estimation was evaluated separately in implant-free and implant-containing samples (Table 2). In implant-free internal radiographs (n = 143), the baseline pipeline showed reliable performance, with ICC values exceeding 0.80 for all parameters. In implant-containing internal radiographs (n = 91), RSM improved parameter estimation most prominently for the L4–L5 SCA, while PT, LL, and SS showed smaller but generally favorable changes. The mean L4–L5 SCA error decreased to 4.7° for both raters from 16.2° (Rater 1) and 15.6° (Rater 2), corresponding to an average reduction of approximately 70\%. The ICC for L4–L5 SCA improved to 0.54 and 0.59 from 0.18, indicating an improvement from poor to moderate reliability. ICCs for PT, LL, and SS were generally above 0.70. On the external test set, L4–L5 SCA error was lower with RSM than with the baseline model, while PT, LL, and SS showed slightly improved or comparable performance. 

Inter- and intra-rater reliability are summarized in Supplementary Tables 4 and 5, respectively. 
Inter-rater agreement, measured using ICC(A,1), was excellent (0.941--0.988) for the global spinopelvic parameters and good (0.871) for the L4–L5 SCA. Mean differences and absolute differences were small overall (-0.19° -- 1.23° and 0.68° -- 1.64°, respectively), with slightly wider variability observed for L4–L5 SCA. The 95\% limits of agreement (LoA) were narrow for most parameters, although wider variability was observed for L4–L5 SCA. Intra-rater ICC(3,1) values were also excellent (0.956 -- 0.999) for all parameters. Differences between repeated measurements were minimal (-0.22° -- 0.81° and 0.36° -- 1.76°, respectively), indicating high repeatability and negligible measurement bias.

Pairwise comparisons of per-image absolute errors between the baseline and RSM pipelines are summarized in Supplementary Table 10. After Benjamini--Hochberg correction, L4–L5 SCA error was significantly lower with RSM in the internal cohort ($q = 1.8 \times 10^{-8}$ and $2.2 \times 10^{-9}$ for Raters~1 and~2), whereas PT, SS, and LL showed no significant differences. In the external cohort, the BH-adjusted q-value (0.077) for the L4–L5 SCA was slightly above the significance threshold (0.05), indicating that the effect of ARNAI was suggestive but not formally statistically significant after multiple-comparison correction.

Additional experimental results are provided in the Supplementary Material, including analyses across segmentation models, high-density out-of-distribution objects, acquisition postures, anatomy preservation, model ablations, preprocessing hyperparameters, and BH-FDR-corrected statistical tests.

\section{Discussion}
\label{sec:discussion}
This study demonstrates the feasibility of the RSM framework for automated spinopelvic measurement in implant-containing lateral lumbar radiographs, where conventional deep learning methods often show reduced performance because of metallic artifacts. By incorporating ARNAI as a preprocessing step, RSM improved delineation of anatomical boundaries obscured by implants, increasing the mean DSC to 0.8697 from 0.8136 and reducing the L4–L5 SCA error by approximately 70\%. These improvements are clinically meaningful because they enhance the robustness of anatomical landmark detection required for automated measurement rather than merely improving image appearance.

The proposed approach is clinically relevant because postoperative lateral lumbar radiographs remain essential for routine follow-up after lumbar fusion surgery, yet implant-related visual interference frequently hinders automatic measurement. Previous approaches often excluded implant-containing radiographs or required manually annotated implant datasets. In contrast, ARNAI was trained exclusively on implant-free images and does not require implant-mask annotation, potentially improving practicality and scalability in clinical workflows.

The effect of ARNAI should not be interpreted as literal restoration of hidden vertebral anatomy. Instead, the method likely improves contour continuity and stabilizes downstream landmark inference. This interpretation is supported by the substantial reduction in L4–L5 SCA error and the high reproducibility of the external reference measurements. Inter-rater ICC(A,1) values ranged from 0.871 to 0.988, whereas intra-rater ICC(3,1) values ranged from 0.956 to 0.999. Inter-rater agreement was relatively lower for the L4–L5 SCA than for global spinopelvic parameters, suggesting that local segmental measurements remain more sensitive to endplate definition and local image quality. Overall, the findings support the role of ARNAI in improving postoperative measurement robustness under implant-related distortion.

Several limitations should be acknowledged. First, the external cohort was relatively small and derived from a single tertiary referral hospital. Second, the subset used for external reference-based evaluation included only radiographs with adequate visualization of key landmarks, which may limit generalizability to more challenging examinations. Third, our experiments focused on the L4–L5 level and were not designed to evaluate all postoperative constructs or surgical levels.
In addition, RSM slightly reduced DSC values for the sacrum, F1, and F2 (Table 1). Because the framework was primarily designed to reduce implant-related artifacts around operated lumbar levels, its benefits were greater for local lumbar measurements than for pelvic landmark detection. While the decreases in DSC values were minimal, they may nevertheless indicate slightly reduced anatomy-preservation performance in excessively bright pelvic regions.
Future studies should therefore include larger multicenter cohorts, broader implant configurations, and validation in other imaging modalities such as CT and MRI.
Despite these limitations, the present findings support the feasibility of a generative artifact-removal framework for automated postoperative spinopelvic analysis without requiring labeled implant-containing radiographs for training.

\printbibliography
\clearpage
%TC:ignore

\begin{table*}[!htbp]
\centering
\caption{Segmentation performance of RSM in terms of DSC, compared with FCBFormer. Models were trained solely on the implant-free internal trainset and evaluated on both the internal test set and an external test set to demonstrate generalization capability. Values are reported as mean (standard deviation).}
\label{tab:before_after_ARNAI}
\resizebox{\textwidth}{!}{%
\begin{tabular}{lccccccccc}
\toprule
\textbf{Model} & \textbf{L1} & \textbf{L2} & \textbf{L3} & \textbf{L4} & \textbf{L5} & \textbf{Sacrum} & \textbf{F1} & \textbf{F2} & \textbf{Mean} \\ \midrule
\multicolumn{10}{l}{\textit{\textbf{Internal test set}}} \\ \midrule
FCBFormer \cite{FCBFormer} & 0.8395 (0.2070) & 0.8673 (0.1966) & 0.7852 (0.2406) & 0.6620 (0.1890) & 0.6807 (0.1950) & \textbf{0.8737 (0.0515)} & \textbf{0.9069 (0.0681)} & 0.8932 (0.0806) & 0.8136 (0.1087) \\
RSM (Ours) & \textbf{0.8575 (0.1781)} & \textbf{0.8860 (0.1610)} & \textbf{0.8811 (0.1613)} & \textbf{0.8461 (0.1594)} & \textbf{0.8170 (0.1771)} & 0.8733 (0.0498) & 0.9033 (0.0718) & \textbf{0.8935 (0.0785)} & \textbf{0.8697 (0.0970)} \\ \midrule
\multicolumn{10}{l}{\textit{\textbf{External test set (Generalization)}}} \\ \midrule
FCBFormer \cite{FCBFormer} & 0.7905 (0.2669) & 0.8480 (0.2471) & 0.8332 (0.2237) & 0.6737 (0.1773) & 0.6661 (0.1897) & \textbf{0.8951 (0.0375)} & \textbf{0.8930 (0.0944)} & \textbf{0.8735 (0.1331)} & 0.8091 (0.1712) \\
RSM (Ours) & \textbf{0.7992 (0.2433)} & \textbf{0.8642 (0.1996)} & \textbf{0.8789 (0.1599)} & \textbf{0.8414 (0.1349)} & \textbf{0.8109 (0.1441)} & 0.8771 (0.0495) & 0.8803 (0.0887) & 0.8650 (0.1345) & \textbf{0.8521 (0.1443)} \\ \bottomrule
\end{tabular}%
}
\end{table*}

\begin{table}[htbp]
\centering
\caption{Comprehensive statistical evaluation of spinopelvic parameter estimation on the internal test set (143 implant-free images and 91 implant-containing images) and the external test set (88 implant-containing images). Only radiographs accepted by both models were included in the analysis.}
\label{tab:comprehensive_spinopelvic}
\setlength{\tabcolsep}{3pt} % 열 사이의 간격을 좁게 설정 (기본값은 보통 6pt)
% \resizebox{가로길이}{세로길이}{내용}
% \textwidth는 페이지 전체 너비, !는 가로 비율에 맞춰 세로를 자동 조절함을 의미
\resizebox{\textwidth}{!}{
    \begin{tabular}{llcccccccc}
    \toprule
    \multirow{2}{*}{\textbf{Param.}} & \multirow{2}{*}{\textbf{Cohort / Model}} & \multicolumn{4}{c}{\textbf{Relative to Rater 1}} & \multicolumn{4}{c}{\textbf{Relative to Rater 2}} \\
    \cmidrule(lr){3-6} \cmidrule(lr){7-10}
    & & \textbf{Mean (SD)} & \textbf{Median (IQR)} & \textbf{R} & \textbf{ICC (95\% CI)} & \textbf{Mean (SD)} & \textbf{Median (IQR)} & \textbf{R} & \textbf{ICC (95\% CI)} \\
    \midrule
    \multicolumn{10}{l}{\textbf{Internal test set}} \\
    \midrule
    \multirow{3}{*}{PT} & Implant-free & 1.7 (3.1) & 1.1 (1.2) & 0.93 & 0.92 (0.88--0.94) & 1.1 (1.4) & 0.9 (1.1) & 0.98 & 0.98 (0.97--0.99) \\
    & Implant (Baseline) & 2.3 (3.9) & 1.2 (1.9) & 0.87 & 0.87 (0.81--0.91) & 2.4 (4.2) & 1.3 (1.8) & 0.86 & 0.85 (0.78--0.90) \\
    & Implant (RSM) & \textbf{2.0 (3.8)} & \textbf{0.9 (1.4)} & \textbf{0.88} & \textbf{0.88 (0.83--0.92)} & \textbf{2.3 (3.9)} & \textbf{1.2 (1.9)} & \textbf{0.87} & \textbf{0.87 (0.81--0.91)} \\
    \midrule
    \multirow{3}{*}{LL} & Implant-free & 4.6 (3.9) & 3.4 (5.3) & 0.92 & 0.88 (0.72--0.94) & 5.0 (3.6) & 4.1 (4.6) & 0.92 & 0.88 (0.64--0.94) \\
    & Implant (Baseline) & 7.8 (6.2) & 6.5 (7.7) & 0.77 & 0.71 (0.47--0.83) & 7.6 (6.4) & 6.1 (7.8) & 0.77 & 0.72 (0.50--0.84) \\
    & Implant (RSM) & \textbf{7.4 (5.6)} & \textbf{6.2 (7.3)} & \textbf{0.79} & \textbf{0.74 (0.52--0.85)} & \textbf{7.1 (5.9)} & \textbf{5.8 (7.1)} & \textbf{0.79} & \textbf{0.75 (0.55--0.85)} \\
    \midrule
    \multirow{3}{*}{SS} & Implant-free & 3.8 (3.1) & 2.9 (4.0) & 0.86 & 0.81 (0.60--0.90) & 4.3 (3.0) & 3.6 (3.6) & 0.87 & 0.80 (0.46--0.90) \\
    & Implant (Baseline) & 7.2 (5.8) & 5.8 (6.0) & 0.67 & 0.61 (0.35--0.76) & 6.8 (6.4) & 4.7 (7.4) & 0.64 & 0.61 (0.44--0.74) \\
    & Implant (RSM) & \textbf{6.1 (4.2)} & \textbf{5.3 (5.7)} & \textbf{0.81} & \textbf{0.73 (0.34--0.87)} & \textbf{5.6 (4.6)} & \textbf{4.7 (4.2)} & \textbf{0.8} & \textbf{0.75 (0.55--0.86)} \\
    \midrule
    \multirow{3}{*}{L4–L5 SCA} & Implant-free & 2.8 (2.5) & 2.1 (3.0) & 0.84 & 0.82 (0.76--0.87) & 3.1 (2.4) & 2.5 (3.4) & 0.82 & 0.81 (0.74--0.86) \\
    & Implant (Baseline) & 16.2 (18.9) & 11.5 (15.1) & 0.44 & 0.18 (--0.01--0.37) & 15.6 (19.3) & 11.1 (16.4) & 0.4 & 0.18 (--0.02--0.36) \\
    & Implant (RSM) & \textbf{4.7 (4.8)} & \textbf{3.1 (5.8)} & \textbf{0.56} & \textbf{0.54 (0.37--0.67)} & \textbf{4.7 (4.3)} & \textbf{2.9 (5.0)} & \textbf{0.6} & \textbf{0.59 (0.44--0.71)} \\
    \midrule
    \multicolumn{10}{l}{\textbf{External test set}} \\
    \midrule
    \multirow{2}{*}{PT} & Implant (Baseline) & 1.6 (3.2) & \textbf{0.8 (1.2)} & 0.88 & 0.88 (0.82--0.92) & 1.6 (2.9) & 1.0 (1.4) & 0.90 & 0.90 (0.85--0.93) \\
    & Implant (RSM) & \textbf{1.5 (2.0)} & 0.9 (1.3) & \textbf{0.94} & \textbf{0.94 (0.91--0.96)} & \textbf{1.5 (1.3)} & 1.0 (1.5) & \textbf{0.97} & \textbf{0.96 (0.94--0.98)} \\
    \midrule
    \multirow{2}{*}{LL} & Implant (Baseline) & 9.8 (9.0) & \textbf{7.3 (7.9)} & 0.66 & 0.51 (0.10--0.73) & 9.9 (9.1) & \textbf{7.6 (7.7)} & 0.65 & 0.50 (0.11--0.72) \\
    & Implant (RSM) & \textbf{9.0 (6.0)} & 8.4 (8.9) & \textbf{0.78} & \textbf{0.65 (0.24--0.82)} & \textbf{9.3 (6.0)} & 8.3 (9.3) & \textbf{0.76} & \textbf{0.64 (0.24--0.81)} \\
    \midrule
    \multirow{2}{*}{SS} & Implant (Baseline) & 7.1 (7.5) & \textbf{5.5 (5.0)} & 0.59 & 0.48 (0.14--0.68) & 6.3 (7.5) & \textbf{4.6 (5.5)} & 0.57 & 0.49 (0.23--0.66) \\
    & Implant (RSM) & \textbf{6.8 (4.4)} & 6.5 (7.3) & \textbf{0.81} & \textbf{0.67 (0.13--0.85)} & 6.3 (4.4) & 5.6 (6.8) & \textbf{0.77} & \textbf{0.67 (0.33--0.82)} \\
    \midrule
    \multirow{2}{*}{L4–L5 SCA} & Implant (Baseline) & 14.2 (12.1) & 11.3 (17.7) & 0.04 & 0.02 (--0.12--0.18) & 14.5 (12.3) & 11.2 (18.4) & 0.04 & 0.02 (--0.11--0.17) \\
    & Implant (RSM) & \textbf{9.5 (10.8)} & \textbf{4.1 (11.9)} & --0.04 & --0.03 (--0.24--0.18) & \textbf{9.7 (10.7)} & \textbf{4.9 (11.5)} & --0.05 & --0.04 (--0.24--0.18) \\
    \bottomrule
    \end{tabular}
}
\par
\vspace{2mm}
\parbox{\textwidth}{
    \scriptsize
    Note: \textit{Implant-free} denotes preoperative radiographs without spinal implants. \textit{Implant} denotes postoperative radiographs containing spinal implants. \textit{Baseline} refers to the baseline pipeline (FCBFormer only), whereas \textit{RSM} refers to the proposed framework (ARNAI + FCBFormer). SD, IQR, and \textit{R} denote standard deviation, interquartile range, and Pearson’s r, respectively. \textit{PT} pelvic tilt, \textit{LL} lumbar lordosis, \textit{SS} sacral slope, \textit{SCA} segmental Cobb angle, CI confidence interval.
}

\end{table}

\clearpage

%TC:endignore

\end{document}